\pdfoutput=1

\documentclass[11pt]{article}

\usepackage[final]{acl}

\usepackage{times}
\usepackage{latexsym}
\usepackage{longtable}
\usepackage[notransparent]{svg}

\usepackage[T1]{fontenc}

\usepackage[utf8]{inputenc}

\usepackage{microtype}

\usepackage{inconsolata}

\usepackage{graphicx}

\title{Datasets for Depression Modeling in Social Media: An Overview}

\author{Ana-Maria Bucur\textsuperscript{1,2}, Andreea-Codrina Moldovan\textsuperscript{1}, Krutika Parvatikar\textsuperscript{3}\\ \vspace{1.5mm} 
\textbf{Marcos Zampieri}\textsuperscript{4}, \textbf{Ashiqur R. KhudaBukhsh}\textsuperscript{3}, \textbf{Liviu P Dinu}\textsuperscript{5,6} \\
\textsuperscript{1}Interdisciplinary School of Doctoral Studies, University of Bucharest, Romania\\
\textsuperscript{2}PRHLT Research Center, Universitat Politècnica de València, Spain\\
\textsuperscript{3}Rochester Institute of Technology, USA, \textsuperscript{4}George Mason University, USA\\
\textsuperscript{5}Faculty of Mathematics and Computer Science, \textsuperscript{6}HLT Research Center,\\
University of Bucharest, Romania\vspace{1.5mm}\\
  \texttt{ana-maria.bucur@drd.unibuc.ro}
}

\begin{document}
\maketitle
\begin{abstract}

Depression is the most common mental health disorder, and its prevalence increased during the COVID-19 pandemic. As one of the most extensively researched psychological conditions, recent research has increasingly focused on leveraging social media data to enhance traditional methods of depression screening. This paper addresses the growing interest in interdisciplinary research on depression, and aims to support early-career researchers by providing a comprehensive and up-to-date list of datasets for analyzing and predicting depression through social media data. We present an overview of datasets published between 2019 and 2024. We also make the comprehensive list of datasets available online as a continuously updated resource, with the hope that it will facilitate further interdisciplinary research into the linguistic expressions of depression on social media.

\end{abstract}

\section{Introduction}
Depression is the most common mental health disorder, and its prevalence has increased further during the COVID-19 pandemic \cite{wolohan2020estimating,kaseb2022analysis,bucur2025state}. Depression is also one of the most extensively researched mental health disorders in the field of psychology \cite{xu2021scientometrics}. Since the past decade, interdisciplinary researchers have explored this widespread mental disorder using data from social media \cite{de2013predicting,yates2017depression,orabi2018deep,aragon2019detecting,fine2020assessing,uban2021emotion,nguyen2022improving,wang2024explainable,raihan2024mentalhelp,abdelkadir2024diverse}. The language used on social media has been shown to predict future depression diagnoses recorded in medical files, suggesting that social media data could be a valuable supplement to traditional depression screening methods \cite{eichstaedt2018facebook}.

Interdisciplinary research has gained popularity through workshops and shared tasks focused on computational approaches for analyzing mental disorders, including CLPsych \cite{chim2024overview}, LT-EDI \cite{kayalvizhi2023overview}, eRisk \cite{parapar2024overview}, and MentalRiskES \cite{marmol2023overview}. As the research community shows increasing interest in examining how depression is expressed in social media language, we aim to support early-career researchers and anyone interested in this field by providing a comprehensive list of datasets for analyzing or predicting depression using social media data. Our motivation stems from recent changes in the terms of service and API rate limits for popular social media platforms, such as Twitter and Reddit, which have been the primary sources for data collection \cite{harrigian2021state}. These changes have made it more challenging and costly to gather new data. Therefore, we focus on the availability of the datasets in this overview. 

The most recent review of social media data for mental health research was conducted by \citet{harrigian2021state}, which covered datasets published between 2014 and 2019. Our current work aims to provide an updated overview of social media datasets specifically related to depression research. Since the latest dataset included by \citet{harrigian2021state} is from 2019, our focus will be on datasets published between 2019 and 2024.

This paper contributes to the computational research in depression by providing a meticulously curated, up-to-date, and continuously updated list of data collections.\footnote{We make the list available online at \href{https://github.com/bucuram/depression-datasets-nlp}{https://github.com/bucuram/depression-datasets-nlp}.} We hope that the resources presented in this overview will further contribute to the interdisciplinary research on depression manifestations in social media language and aid in developing effective interventions for those affected by depression.

\section{Methodology}
We have conducted a comprehensive literature search on the major publication databases, including ACL Anthology, IEEE Xplore, Scopus, ACM Digital Library, Springer Nature Link, ScienceDirect, and Google Scholar to search for papers using NLP models for depression modeling or papers presenting novel depression-related data collections from social media. We formulated the following search query to retrieve relevant papers:

\noindent (“depression” OR “depression detection” OR “depression prediction” OR “depression monitoring” OR “depression analysis”) AND (“social media” OR “online” OR “Twitter” OR “Reddit” OR “Facebook”)

For this overview, we selected papers published between 2019 and 2024 that specifically analyze depression using social media data. We excluded any papers not written in English. To determine if the retrieved papers included analyses related to depression based on social media data or described new data collections, we manually inspected the full texts. We focused on data in the English language. In total, we identified 310 relevant papers, of which 59 proposed new data collections for depression-related research using social media data.

\section{Datasets}
\label{sec:data}
\begin{figure}[!ht]
	\centering
		\includegraphics[scale=.5]{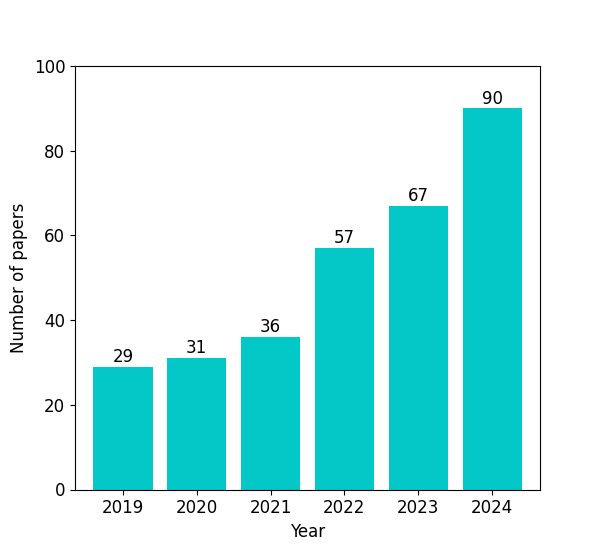}
	\caption{Number of papers on depression modeling published each year in peer-reviewed conferences or journals.\label{num-of-papers-depression}}
\end{figure}

In Figure \ref{num-of-papers-depression}, we show the number of papers on depression modeling from social media data published each year. 

We observe a growing trend in interdisciplinary research on depression, which may have been partly influenced by the COVID-19 pandemic, as there has been an increase in depression rates during this time \cite{wolohan2020estimating,kaseb2022analysis}. In addition, there has been more research focused on using NLP models for mental health surveillance on social media platforms to assess the pandemic's impact on the population \cite{dhelim2023detecting}.

\begin{figure}[!ht]
	\centering
		\includegraphics[scale=.44]{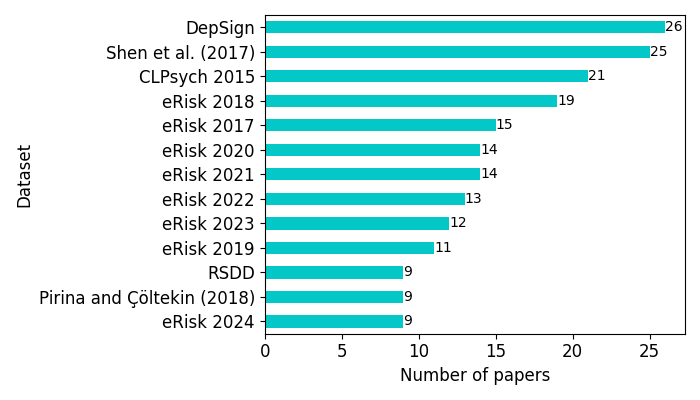}
	\caption[The most used datasets for depression modeling]{The most used datasets for depression modeling.}
	\label{most-used-datasets}
\end{figure}

In Figure \ref{most-used-datasets}, we present the most used datasets in the 310 papers found through our search. Most of the papers have used the datasets from the LT-EDI Workshop (DepSign dataset \cite{Sampath-dataset}), the eRisk Lab \cite{losada2017clef,Losada2018OverviewOE,losada2019overview,losadaoverview,parapar2021overview,crestani2022early}, or the CLPsych 2015 Shared Task dataset \cite{coppersmith2015clpsych}. All the aforementioned datasets were released as part of shared tasks or competitions, and the data was a valuable resource that was further used after the end of the shared task. Other benchmark datasets are from \citet{shen2017depression}, \citet{pirina-coltekin-2018-identifying}, or RSDD \cite{yates2017depression}. 

\noindent The availability of data collections has advanced the development of state-of-the-art depression prediction models. Of the 310 papers published during 2019 and 2024, 59 of them collect and annotate new data from online platforms. In Appendix \ref{sec:appendix} Table \ref{tab:datasets-survey}, we present detailed information for each of the data collections, such as the platform used for data gathering, the annotation procedure, and the level of annotation (either for each post or user), the labels that are provided for the data, the size of the dataset and its availability.

\begin{figure}[!ht]
	\centering
		\includegraphics[scale=0.58]{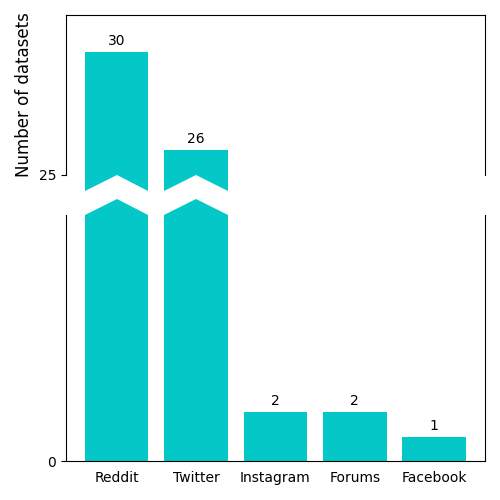}
	\caption{The most used platforms for the data collections presented in this overview.}
	\label{most-used-platforms}
\end{figure}

\paragraph{Platform} In Figure \ref{most-used-platforms}, we present the social media platforms used for gathering datasets for depression modeling. Reddit and Twitter were the most commonly used platforms for data collection due to easy access to dedicated APIs. However, recent changes in the terms of service and API rate limits for both Twitter / X\footnote{\href{https://developer.twitter.com/en/docs/twitter-api/rate-limits}{https://developer.twitter.com/en/docs/twitter-api/rate-limits}} and Reddit\footnote{\href{https://support.reddithelp.com/hc/en-us/articles/16160319875092-Reddit-Data-API-Wiki}{https://support.reddithelp.com/hc/en-us/articles/16160319875092-Reddit-Data-API-Wiki}} have complicated data collection from these platforms. These updates may hinder the reproduction of datasets where authors only provide Twitter or Reddit IDs instead of the raw text. In addition, these changes make the process of collecting new data more challenging, costly and time-consuming.

\begin{figure}[!t]
	\centering
		\includegraphics[scale=0.7]{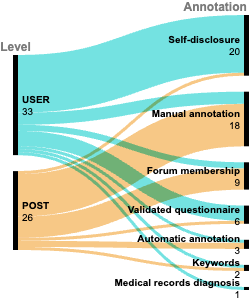}
	\caption{Overview of the annotation levels within each dataset, at either the user or post level, along with the procedures used for annotation.}
	\label{fig:annotation}
\end{figure}

\paragraph{Annotation procedure and labels} For depression detection from social media data, the most common method of annotation from the datasets presented in this work is the annotation based on self-disclosure (Figure \ref{fig:annotation}), labeling users binary, depending on whether they mention online a depression diagnosis or not. In 20 of the data collections, researchers use self-mentions of depression diagnoses (e.g., ``I was diagnosed with depression'') for their annotation processes. This approach allows for the compilation of large datasets containing hundreds of thousands of users.

Another common annotation procedure is manual annotation, used for 18 of the data collections. These annotations can be performed by mental health experts, graduate students, or laypeople. Most procedures for manual annotations are performed at the post level. Manual annotation is used to label the data binary (depression vs. control), to label data for depression severity (no signs of depression, mild, moderate, severe, etc.), and for symptoms measured by different validated questionnaires, or symptoms from The Diagnostic and Statistical Manual of Mental Disorders, Fifth Edition (DSM-V) \cite{APA2013}. Recently, datasets have shifted from binary labeling to labeling based on depression symptoms, leading to the development of explainable methods for depression modeling \cite{anxo2023psyprof,bao2024explainable}.

Data annotation can also be performed by asking social media users to fill in validated self-report questionnaires, such as the Beck's Depression Inventory (BDI) or Patient Health Questionnaire-9 (PHQ-9). However, even if psychometric tools produce a more reliable assessment of depression, fewer people are willing to participate in the data collection, resulting in small sample sizes. Only six datasets rely on self-report questionnaires for the annotation procedure, and one of them relies on the diagnosis from medical records.

Another method for annotation, which is noisier and more prone to errors, is labeling posts by the presence of specific depression-related keywords or automatic annotation performed via an NLP model trained on mental health data. These methods are used less frequently in the data collections included in this overview, with only three data collections being labeled automatically and two datasets being labeled using depression-related keywords.

\paragraph{Availability} Due to the sensitive nature of the information in the datasets used for depression modeling, their availability varies. Our exploration of data availability was inspired by the work of \citet{harrigian2021state}. However, unlike their study, we have decided not to consider datasets that can be reproduced using APIs from social media platforms as readily available. This decision was influenced by recent changes in the terms of service of platforms such as Reddit and Twitter / X, which have complicated the reproduction of data and made it difficult to retrieve social media posts using the IDs included in the data collections via APIs.

Out of the 59 papers proposing new datasets, 16 are publicly available and hosted online for anyone to use, 15 can be made available after signing a data usage agreement, and 11 collections can be made available by contacting the authors of the dataset. The availability of the rest of the datasets is unknown.

\section{Discussion}
\paragraph{Data availability} One of the primary motivations for this overview were the recent changes in social media platforms, which may hinder the development of new research collections. Our aim was to provide the research community with a comprehensive list of data collections that can be used for interdisciplinary research on the manifestations of depression in social media. We included availability information for each dataset in this overview. We have found that 16 of the datasets are publicly available and free for anyone interested to download and use. As detailed in Section \ref{sec:data}, data collections that were part of shared tasks or easily accessible were successfully used by the research community.

\paragraph{Annotation reliability} 
One common method for user-level labeling involves relying on individuals to self-disclose their depression diagnoses. However, this approach is not reliable. Even when annotators manually review posts that contain self-disclosed information, there is no way to verify the authenticity of these disclosures or the accuracy of the users' statements. In addition, for the control group, which includes users who do not mention any depression diagnoses, their actual mental health status remains unknown. We cannot assume that these individuals do not suffer from mental disorders because they have not disclosed this information. It is essential to recognize that relying on self-reported diagnoses for mental health data collection can lead to self-selection bias  \cite{amir-etal-2019-mental}. This means that the data obtained may only represent individuals who are willing to openly discuss their mental health issues, which may not accurately reflect the entire population of people with mental disorders.

\section{Conclusion and Future Work}
We presented a comprehensive and up-to-date overview of datasets used for depression modeling from social media data. We review papers published in international conferences and journals between 2019 and 2024. Due to the research community's efforts to organize shared tasks, the availability of benchmark datasets has increased, offering researchers the resources to build online screening methods for depression and to analyze the depression-related discourse online. 

This paper not only aims to offer information about the available datasets for depression manifestation in social media language, but to encourage further interdisciplinary collaboration and exploration. We hope that the comprehensive list of resources provided will inspire researchers, particularly those in the early stages of their careers, to explore this field more deeply. This could lead to a better understanding of depression as expressed in social media and improved interventions.

In this overview, we focused on English datasets, as it is one of the languages that are most used for data collection \cite{harrigian2021state,skaik2020using}. However, studying the manifestations of mental health problems in low-resourced languages is an important step toward providing depression screening solutions that can improve the mental health outcomes of people from all around the world \cite{garg2024towards}. In future work, we aim to extend this effort to include social media datasets in languages other than English. Furthermore, we would like to explore the relationship between datasets curated for depression detection and those used in related tasks. This would provide insights on the relationship between depression detection and related social media tasks \cite{bucur2021exploratory} as well as support multi-task learning efforts \cite{benton-etal-2017-multitask,kodati2025advancing}.

\section*{Limitations}
In this paper, we aim to provide a comprehensive overview of the current state of social media data for computational research on depression and present a list of datasets available for researchers in this field. Our study includes 59 data collections, each of which has been carefully reviewed. However, it is possible that we may have overlooked some works that do not explicitly mention depression-related analyses using social media data in their titles or abstracts.

\section*{Ethical Considerations}
Addressing ethical considerations in mental health research that uses social media data is essential for protecting the privacy, confidentiality, and well-being of individuals whose data is being analyzed \cite{chancellor2020methods,benton2017ethical,chancellor2019taxonomy}. In this overview, we present the datasets available for studying the manifestations of mental disorders on social media. Although we do not conduct any analyses on the data presented in this work, we want to emphasize that collecting social media data from individuals affected by mental disorders must adhere to ethical research protocols \cite{benton2017ethical}. Additionally, researchers who use these datasets should follow the same ethical guidelines and recommendations for health research involving social media.

\section*{Acknowledgements}

This research was partially supported by the project “Romanian Hub for Artificial Intelligence - HRIA”, Smart Growth, Digitization and Financial Instruments Program, 2021-2027, MySMIS no. 334906.

\bibliography{custom}

\clearpage
\onecolumn

\section{Appendix}
\label{sec:appendix}

    {\footnotesize
    \begin{longtable}[c]{p{4cm}|p{1cm}|p{0.7cm}|p{3cm}|p{2cm}|p{1.5cm}|p{1cm}}
    \captionsetup{singlelinecheck = false}
    
    \caption{\small List of available datasets for depression modeling using data posted on online platforms. The labels for availability are the following: \textbf{FREE} - the dataset is publicly available and hosted online for anyone to access, \textbf{AUTH} - the data can be accessed by contacting the paper's authors, \textbf{DUA} - the data is available only after a data usage agreement is signed, \textbf{UNK} - the dataset availability is unknown; the authors do not mention if the data is available to the research community.
    }\label{tab:datasets-survey}\\
    \hline
    \textbf{Dataset}& \textbf{Platform}& \textbf{Level}& \textbf{Annotation Procedure} & \textbf{Label} & \textbf{Size} & \textbf{Availab.}\\
    \hline
    \endfirsthead
    
    \textbf{Dataset}& \textbf{Platform}& \textbf{Level}& \textbf{Annotation Procedure} & \textbf{Label} & \textbf{Size} & \textbf{Availab.}\\
    \hline
    \endhead
    
    \hline
    \endfoot
    
    \hline
    \endlastfoot

\citet{gui2019cooperative} & Twitter  & USER  & Self-disclosure  & Binary  & 2.8K users  & UNK\\
\citet{Chandra_Guntuku_Preotiuc-Pietro_Eichstaedt_Ungar_2019} & Twitter  & USER  & BDI  & Binary  & 887 users  & UNK\\
\citet{almouzini2019detecting} & Twitter  & USER, POST & Manual annotation  & Binary  & 89 users  & UNK\\
eRisk2019 \cite{losada2019overview}  & Reddit  & USER  & BDI-II  & BDI filled-in  & 20 users  & DUA\\
\citet{owen2020preemptive}  & Twitter  & POST  & Manual annotation  & Binary  & 1K posts  & FREE\\
\citet{DBLP:journals/corr/abs-2002-02800} & Twitter  & USER  & Self-disclosure  & Binary  & 1.2K users  & AUTH\\
\citet{rissola2020dataset} & Reddit  & POST  & Self-disclosure, heuristics  & Binary  & 14K posts  & DUA\\
\citet{articlebirn}  & Facebook  & USER  & Medical records diagnosis  & Binary  & 223 users  & AUTH\\
D2S \cite{DBLP:journals/corr/abs-2011-06149} & Twitter  & POST  & PHQ-9  & PHQ-9 symptoms  & 12K posts  & AUTH\\
eRisk 2020 \cite{losadaoverview} & Reddit  & USER  & BDI-II  & BDI filled-in  & 90 users  & DUA\\
\citet{tabak2020temporal} & Twitter  & USER  & Self-disclosure  & Binary  & 5K users  & UNK\\
\citet{Yazdavar2020MultimodalMH} & Twitter  & USER  & Manual annotation  & Binary  & 8.7K users  & DUA\\
\citet{haque2021deep} & Reddit & POST  & Subreddit participation  & Depression vs. suicide  & 1.8K posts  & FREE\\
\citet{chiu2021multimodal} & Instagram   & USER  & Depression-related keywords  & Binary  & 520 users  & UNK\\
\citet{nanomi-arachchige-etal-2021-dataset-research} & Online forums   & POST  & Manual annotation  & Depression severity   & 2.1K posts  & UNK\\
\citet{sherman-etal-2021-towards} & Reddit  & USER  & Self-disclosure  & Binary  & 31K users  & DUA\\
eRisk 2021 \cite{parapar2021overview} & Reddit  & USER  & BDI-II  & BDI filled-in  & 170 users  & DUA\\
\citet{10.1145/3459637.3482366} & Twitter  & USER  & Self-disclosure  & Binary  & 817 users  & AUTH\\
\citet{guo2021emotion}& Reddit  & USER  & Self-disclosure  & Labels for multiple disorders  & 7.9 K users  & UNK\\
\citet{zhang2021monitoring}  & Twitter  & USER  & Self-disclosure  & Binary  & 5K users  & UNK\\
\citet{zhou2021detecting} & Twitter  & USER  & Self-disclosure  & Binary  & 1.8M posts  & UNK\\
\citet{articlesafa} & Twitter  & USER  & Self-disclosure  & Binary  & 1.1 K users  & AUTH\\
\citet{10.1145/3485447.3512128} & Reddit  & POST  & Manual annotation  & Depression severity   & 3.5 K posts  & FREE\\
PsySym \cite{zhang-etal-2022-symptom} & Reddit  & USER, POST & Automatic and manual annotation  & DSM-5 symptoms for multiple disorders & 26K users, 8.5K posts & AUTH\\
MHB \cite{boinepelli-etal-2022-leveraging} & Online forums  & USER  & Forum participation   & Only depression  & 9.3K users  & FREE\\
CAMS \cite{garg-etal-2022-cams}& Reddit  & POST  & Manual annotation  & Causes for depression  & 3.1 K posts  & FREE\\
\citet{sotudeh2022mentsum} & Reddit  & POST  & Subreddit participation  & Summarization  & 24 k posts  & DUA\\
\citet{Sampath-dataset}  & Reddit  & POST  & Manual annotation  & Depression severity   & 16K posts  & FREE\\
eRisk2022 \cite{crestani2022early}  & Reddit  & USER  & Self-disclosure  & Binary  & 3.1K users  & DUA\\
\citet{10.1145/3487553.3524918} & Reddit  & POST  & Subreddit participation  & Labels for multiple disorders  & 16 K posts  & UNK\\
PRIMATE \cite{gupta-etal-2022-learning} & Reddit  & POST  & Manual annotation  & PHQ-9 symptoms  & 2K posts  & DUA\\
PsycheNet-G \cite{mihov2022mentalnet} & Twitter  & USER  & Self-disclosure  & Binary  & 591 users  & UNK\\
Twitter-STMHD \cite{singh2022twitter} & Twitter  & USER  & Self-disclosure, manual annotation  & Labels for multiple disorders  & 33K users  & FREE\\
multiRedditDep \cite{uban2022explainability}& Reddit  & USER  & Self-disclosure  & Binary  & 3.7K users  & AUTH\\
\citet{articlesdht} & Reddit  & USER  & Subreddit participation  & Binary  & 81K users  & UNK\\
\citet{FernandezBarrera2022evaluating}  & Flickr  & POST  & Depression tags  & Only depression  & 14.5K posts  & UNK\\
\citet{cha2022lexicon}  & Twitter, Everytime  & POST  & Lexicon-based automatic annotation  & Binary  & 26M posts, 22K posts  & AUTH\\
DEPTWEET \cite{KABIR2023107503}  & Twitter  & POST  & Manual annotation  & Depression severity   & 40K posts  & FREE\\
\citet{ALAVIJEH2023103269} & Twitter  & USER  & Self-disclosure  & Labels for multiple disorders  & 1.5K users  & FREE\\
\citet{ADARSH2023103168} & Reddit  & POST  & Subreddit participation  & Binary  & 60K posts  & UNK\\
\citet{10.1145/3578503.3583621} & Reddit  & POST  & Subreddit participation  & Symptoms  & 1.3M posts  & FREE\\
BDI-Sen \cite{perez2023bdi} & Reddit  & POST  & Manual annotation  & BDI-II symptoms  & 4.9K posts  & DUA\\
\citet{song2023simple} & Reddit  & POST  & Subreddit participation  & Labels for multiple disorders  & 85K posts  & UNK\\
RedditCE \cite{liang2023identifying} & Reddit  & POST  & Manual annotation  & Emotion-cause labels   & 35K posts  & FREE\\
\citet{liu2023improving} & Reddit, Twitter  & USER  & Self-disclosure  & Binary  & 205K users, 255 users & UNK\\
RESTORE \cite{yadav2023towards} & Reddit, Twitter, Pinterest  & POST  & Manual and automatic annotation  & PHQ-9 symptoms  & 9.8K images  & AUTH\\
\citet{zogan2023hierarchical} & Twitter  & USER  & Self-disclosure  & Binary  & 1.4K users  & UNK\\
\citet{wu2023exploring} & Twitter  & USER  & Self-disclosure, manual annotation  & Binary  & 10K users  & DUA\\
DepreSym \cite{perez2023depresym} & Reddit  & POST  & Manual annotation  & BDI-II symptoms  & 21K posts  & DUA\\
\citet{10315126} &  Twitter  &  USER  &  Self-disclosure  & Labels for multiple disorders  &  6K users  &  DUA\\
HelaDepDet \cite{priyadarshana2023heladepdet} &  Twitter, Reddit  &  POST  &  Manual annotation  & Depression severity  &  40K posts  &  FREE\\
\citet{anshul2023multimodal} &  Twitter  &  USER  &  Self-disclosure, Manual annotation  &  Binary  &  1.5K users  &  FREE \\
RED \cite{welivita2023empathetic} & Reddit & POST & Subreddit participation  & Labels for multiple disorders & 1.2M posts & FREE\\
\citet{alhamed2024classifying} &  Twitter  &  USER  &  Manual annotation  &  Before/After diagnosis  &  120 users  &  FREE\\
\citet{milintsevich2024your} &  Reddit  &  POST  &  Manual annotation  &  Anhedonia  &  167 posts  &  DUA\\
MentalHelp \cite{raihan2024mentalhelp} &  Reddit  &  POST  &  Automatic annotation  &  Binary  &  14M posts  &  FREE\\
\citet{lee2024detecting} &  Reddit  &  USER  &  Manual annotation  &  Binary  &  1K users  &  DUA\\
\citet{beniwal2024hybrid} &  Instagram  &  POST  &  Manual annotation  &  Binary  &  10K posts  &  AUTH\\
\citet{tumaliuan2024development} &  Twitter  &  USER  &  PHQ-9  &  Binary  &  72 users  &  AUTH\\

\hline


\end{longtable}
}

\end{document}